\documentclass[manuscript, nonacm]{acmart}
\AtBeginDocument{%
  }

\acmConference[Conference acronym 'XX]{Make sure to enter the correct
  conference title from your rights confirmation emai}{June 03--05,
  2018}{Woodstock, NY}
\acmISBN{978-1-4503-XXXX-X/18/06}

\usepackage{graphicx}
\usepackage{float}
\usepackage{caption,setspace}
\usepackage{balance}
\usepackage{adjustbox}
\usepackage{hyperref}
\usepackage{url}
\usepackage{multirow}
\usepackage{subcaption}
\usepackage[english]{babel}
\usepackage{booktabs}
\usepackage{enumerate}
\usepackage{multicol}
\setcopyright{none} 

\begin{document}

\title{Epistemic Bias as a Means for the Automated Detection of Injustices in Text}
\author{Kenya Andrews}
\email{kandre32@uic.edu}
\affiliation{%
  \institution{University of Illinois at Chicago}
  \city{Chicago}
  \state{IL}
  \country{USA}
}

\author{Lamogha Chiazor}
\affiliation{%
  \institution{IBM Research}
  \city{London}
  \country{UK}
  }
\email{lamogha.chiazor@ibm.com}

\renewcommand{\shortauthors}{Andrews and Chiazor 2025}

\begin{abstract}
  Injustices in text are often subtle since implicit biases or stereotypes frequently operate unconsciously due to the pervasive nature of prejudice in society. This makes automated detection of injustices more challenging which leads to them being often overlooked. We introduce a novel framework that combines knowledge from epistemology to enhance the detection of implicit injustices in text using NLP models to address these complexities and offer explainability. Our empirical study shows how our framework can be applied to effectively detect these injustices. We validate our framework using a human baseline study which mostly agrees with the choice of implicit bias, stereotype, and sentiment. The main feedback from the study was the extended time required to analyze, digest, and decide on each component of our framework. This highlights the importance of our automated framework pipeline that assists users in detecting implicit injustices while offering explainability and reducing time burdens on humans.
\end{abstract}

\maketitle

\section{Introduction}
The most basic duty of the media is knowledge sharing. Yet, the tool necessary for wide-spread knowledge sharing is influence. With this influence, the media is able to shape how one will understand the intricacy of a storyline in a news story with little effort. This often results in the use of \emph{epistemic biases} which involves propositions that are presupposed, entailed, asserted, hedged, or boosted in text \citep{recasens2013linguistic} to erode or assert a person's capacity as a knower, leading to framing issues and implicit injustices within the text. Particularly, some of these word choices in text can lead to e.g., epistemic injustices occurring (i.e., when the information is not based on facts or the truth, but more on subjective beliefs). These injustices can affirm or perpetuate stereotypes concerning the subjects involved. Though it has always been a harsh reality with various ringing consequences, in recent years we have publicly witnessed how the affirmation of stereotypes can lead to physical violence, prejudice, and negative self-image \citep{harrison1999myths, gover2020anti,kuykendall1989improving}. These experiences are harmful and dangerous, explicitly for the victims, but also for all members of our society. 

Therefore, we seek to detect instances of testimonial injustice, character injustice, and framing injustice in text. Testimonial injustice occurs when word choices cast modified believability on a statement due to prejudices (e.g., stereotypes) about the sayer \citep{fricker2007epistemic}. Character assassination is ``the  deliberate  destruction of an individual’s reputation  or  credibility" \citep{icks2019character}. This often leads to character injustice, a term we define as an unjustified attack on a person's character that results in an unfair criticism or inaccurate representation of them (e.g., exaggeration or defamation). According to \citep{entman2007framing} framing bias happens when the use of subjective, one-sided words that reveals the stance of an author occurs. Which means that an individual’s choice from a set of options is influenced more by \emph{how the information is worded}, rather than by the information itself. We recognise that text, such as those found in news content can be positively or negatively framed to influence the narratives. In this paper, we aim to show how word choices that negatively frame a statement, rather than presenting it with factual or neutral terms, can result in the unfair depiction of certain subjects, thereby contributing to them experiencing these injustices. 

With this work, we seek to make room for subjects of text, even in creative writing, to not have their credibility shot or character assumed due to well-known stereotypes which are harmful and unfounded. It can be noted here that toxic or hate speech is often blatant and highly recognisable. Which differs from the objective of our work since here we focus on implicit injustices which are less obvious but could be just as harmful as toxic or hate speech. We pursue this, acknowledging it is achievable to speak about any happening in a factual way that does not seek to cause harm on the subject of the text. Further, we seek to offer explainability towards how and why a writing may cause or promote injustice towards the subject of their text.

Our proposed framework will help with the detection of testimonial, character, and framing injustices. The framework includes a fine-tuned BERT model based on work from \citep{pryzant2020automatically} to automatically tag words associated with epistemic bias from an input text, uses the \citep{kwon2021costar} CO-STAR model and \citep{sap-etal-2020-social} Social Bias Frames (SBF) to generate potential associated stereotypes and the concepts of those stereotypes to the input text, and show when the tagged words (associated with some epistemic bias) or less credibility of a person are correlated with a stereotype which causes injustice. Though we could use examples from various fields (e.g., politics, marketing, medicine, etc...), we will use news media as a use-case throughout this work. Thus, we present the following contributions: (1) We develop a novel framework that uses the results of 3 NLP models to automatically detect character, testimonial, and framing injustices in text, (2) We produce a fine-tuned tagger model to automatically detect epistemic biases, (3) We develop an Interactive User-Interface for journalists and editors to submit text to and receive output and explanations surrounding the potential bias inducing tagged word, (4) We produce empirical evidence showing how epistemic bias can translate to injustices, and (5) We conduct a survey to validate the results of our empirical study and feature selection for fine-tuning the tagger model.

An outcome of this work is to give journalists and editors a tool which will help them easily and quickly notice and avoid  testimonial, character, and framing injustices in their work. This will be accomplished by showing users which words they use that produce epistemic bias, the potential stereotype associated with the tagged words and text, offering the user explainability with the help of the stereotype concepts as defined by \citep{kwon2021costar}, and resources to reference literature on the particular epistemic bias type(s) identified in their input text. 

\section{Background} \label{sec:literature}
Many works have established it is difficult for the common person to identify a biased word in a sentence and establish the need for computational agents to take on this charge. \citet{pryzant2020automatically} show humans have low ability to detect bias and show humans perform worse than their detection model. \citet{recasens2013linguistic} show that the accuracy of Humans annotators on AMT (amazon mechanical turk) was not more than 37.39\% for a single detected biased word. The difficulty arises due to us holding our own biases as facts and lack of education on sentence construction. It has also been observed that when human, expert annotators follow guides that help them detect bias, can more easily find instances of such biases. For example, the editors of the Wiki Neutrality Corpus had to follow the Netural Point of View (NPOV) Guidelines \footnote{\url{https://en.wikipedia.org/wiki/Wikipedia:Neutral\_point\_of\_view}}. 

Great efforts have been put towards identifying potential words in text materials that could encourage epistemic bias \citep{recasens2013linguistic, hube2019neural, pryzant2020automatically}. Following on from the work of~\citet{pryzant2020automatically}, we fine-tuned their tagger model to automatically tag words associated with epistemic bias from an input text. Identifying words which cause epistemic bias is a step towards awareness of social harms. This begs the question, what do we do with our new found knowledge and awareness? What kinds of implications does our use of these words impact society? What communities are affected by these word choices? These are the questions we explore in this through our empirical study and validation study as discussed in Section \ref{sec:methodology}.

Authors \citet{kwon2021costar} have trained a model to detect widely-known stereotypes and the concept of those stereotypes in text materials. We leverage the results of their CO-STAR model and the SBF model \citep{sap-etal-2020-social} to offer some explainability of the word choices by the model. Associating a particular text with a stereotype and the concept of that stereotype is a critical step towards awareness of social harms that might cause character or testimonial injustice to a particular individual or group. Lack of identifying the words in a sentence which imply and promote these stereotypes leaves us with the undirected burden and question of: how can we address these harms? Our solution promotes awareness and accountability to directly highlight words that are likely contributing to such harms (using a fine-tuned tagging model), enabling users to recognize and address potentially harmful language more effectively.

\citet{beach2021testimonial} identify words in text that cause testimonial injustice in medical records of Black patients. Detecting such testimonial injustice is helpful in seeing the unjust realities of our society. They conclude the testimonial injustice that persists in these medical records has a high potential of causing disparity in the quality of health care for Black patients, which correlates with findings that Black patients receive a worse quality of health care~\citep{odonkor2021disparities}. Identifying testimonial injustice in text materials is vital to creating an environment of accountability. For accountability to be applied we must include education, which is often unexplored or left up to the user. Many users do not know where to find resources for such things, thus we provide them in our framework.


\citet{raza2022dbias} developed a pipeline which takes in news articles, detects and masks words that are biased, and suggests words with more neutral text. Whilst the pipeline and library designed by the authors are very good and useful, they however do not consider the linguistic and epistemic bias features as discussed in~\citet{recasens2013linguistic} and used by~\citet{pryzant2020automatically}. Unlike our work discussed in this paper, their pipeline produces no way of distinguishing and highlighting epistemic bias types of any identified potential biased word in the sentence. Also, their framework does not attempt to relate potential stereotypes and stereotype concepts that might be associated with causing injustice to a person or group.

\section{Methodology}\label{sec:methodology}

\begin{figure*}[h]
 \centering
 \includegraphics[scale=0.26]{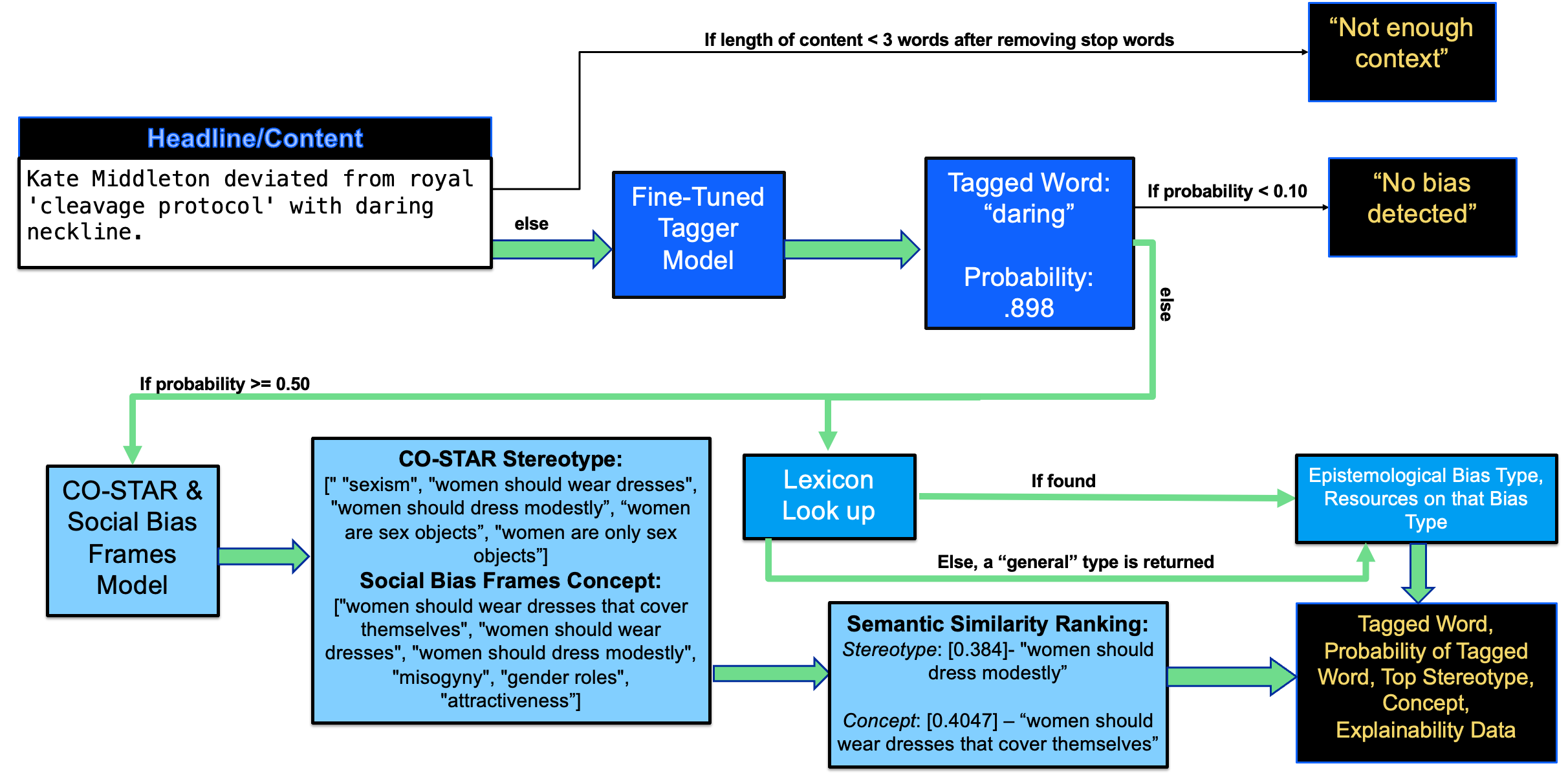}
 \caption{Testimonial Injustice Technical Framework}
 \label{fig:framework}
\end{figure*}

\citet{hamborg2019automated} concluded that various forms of media bias is already analysed in the Social Sciences field and can be implemented in an automated fashion by primarily using Natural Language Processing (NLP) techniques. We leverage research methods using NLP and deep learning whilst also using analysis concepts by researchers from social sciences. Thus, we propose a novel, technical framework (Figure \ref{fig:framework}) which includes using NLP models for detecting potential epistemic biased words and potential stereotypes along with their concepts, which we semantically link to given sentence(s). From the literature, we found very few models which consider linguistic, epistemic features, or common sense reasoning to detect implicit biases or reasoning about potential stereotypes. This in turn influenced our decision to use three baseline models with promising results (i.e., Tagger Model, Co-Star, and SBF). 



\begin{figure*}[h]
 \includegraphics[scale=0.25]{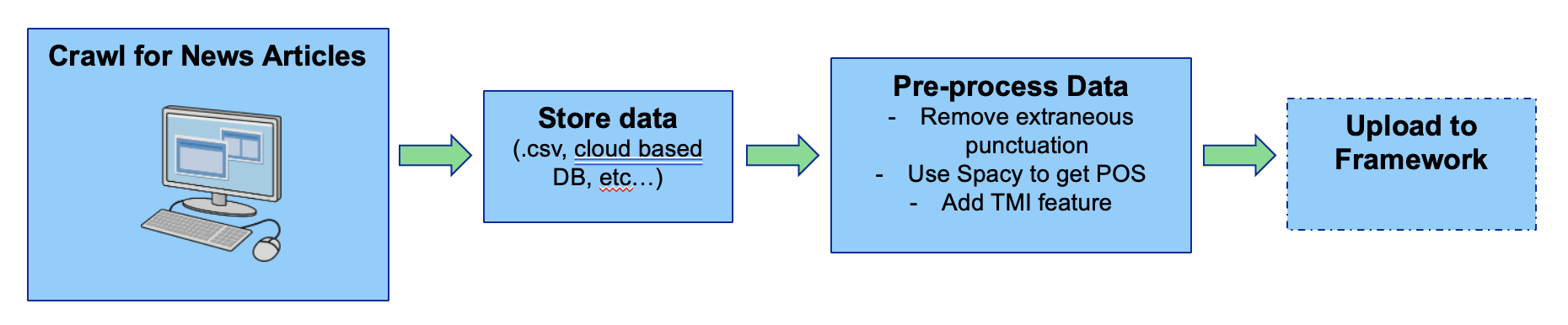}
 \centering
 \caption{Process of Getting Data to the Framework}
 \label{fig:crawling}
\end{figure*}

Figure \ref{fig:crawling}, shows the steps required to scale the framework to analyse multiple headlines. 

\subsection{Tagger model}\label{ssec:methodtagger} 
We leverage the detection component of the modular model as discussed in \citet{pryzant2020automatically} and refer to this as the \emph{tagger model} for the rest of this paper. 

The tagger model (see Figure \ref{fig:tagger-detection-model} in Appendix \ref{sec:appendix}) takes as input sentence(s) (e.g., a headline text, a collection of headlines, etc...), and predicts the probability of each word in the sentence(s) been biased (see Equation \ref{eq:1} and Figure \ref{fig:framework}). Then it returns the word with the highest probability as the tagged biased word. 
\begin{equation}\label{eq:1}
P_i=\sigma (b_iW^b + e_iW^e + b)
\end{equation}

Where \(b_i \in R^b\) is a word's \(w_i^s\) semantic meaning and \(e_i\) are the experts features as proposed by \cite{recasens2013linguistic} and based on Equation \ref{eq:2}.
\begin{equation}\label{eq:2}
   e_i = R_eLU(f_iW^{in}) 
\end{equation}

 The detection model itself is a BERT-based neural sequence tagging model that has been fine-tuned to include the expert linguistic and epistemic bias features from \citet{recasens2013linguistic}. An example of the expert feature as listed in Table 3 of \citet{recasens2013linguistic} is the \textit{part of speech (POS)} tag of each word in the sentence. Another example of an expert feature from vetted experts in linguistics is \textit{“assertive verbs”}. The model leverages the semantic meaning of each word in the given sentence via the BERT \citep{kenton2019bert} contextualised word vector. 

 When there are words in a sentence which do not help with understanding, but are excess information these are known to cloud the judgment of the reader while simultaneously giving the reader reason to be more confident in the conclusions they make \citep{tmilinkedin, spira2011overload}. We hypothesised here that - having \emph{too much information (TMI)} can be considered from an English linguistic standpoint to be a sentence instance which has more than 2 descriptive words (i.e., adjectives and adverbs) in it that do not add to its understanding but seeks to cloud the judgement of the readers. We manually analysed a few biased and non-biased headlines and saw an excessive use of descriptors (i.e., adjectives, adverbs) in headlines that were biased. We extend the detection model further by including a feature on whether or not the sentence contained TMI. We created methods in python which makes use of the core NLP dependency tree and a tree traversal algorithm to go through sentences starting from the root node, then count the number of adjectives and adverbs in the sentence to determine if there is ``TMI" or ``no TMI" based on the hypothesis. TMI is not a known indicator of epistemic bias, but we acknowledge it might contribute to causing doubt which leads to injustice. This is why we introduce this new feature to see if it will be a contributing feature for the tagger model to detect epistemic biases. We conducted an ablation study to see the effects of adding this feature, whilst training the tagger model in Section \ref{sec:abalationstudy}. Further, we conduct a survey to validate our hypothesis about TMI in Section \ref{ssec: valsurveyres}.

\subsection{CO-STAR Model and Social Bias Frames}\label{ssec:methodstereotypes} 

The CO-STAR (COnceptualisation of Stereotypes for Analysis and Reasoning) framework allows input from the user and generates outputs of stereotypes and stereotype concepts using a GPT based model.  The SBF model generates and classifies stereotypes associated with an inputted text. These outputs are quite simple to understand but have a very complex history. The accuracy of the baseline SBF model was analysed by looking for the demographic group the statement targeted and the implied stereotype from the statement. The authors measured the accuracy with BLEU-2 (group - 83.2\%, stereotype - 68.2\%) and Rouge-L (group - 49.9\%, stereotype - 43.5\%). The authors of the CO-STAR model manually evaluated their model, but did not specify the results of their evaluation. However, we analysed how each of these models performed on our sentences and found them to generate stereotypes which are often well fitted to the sentences submitted.
  
Since testimonial injustice occurs because of widely known stereotypes, the outputs of these models will help us inform our users of any potential stereotypes that are exacerbated by their inputted text. The potential stereotype can also be used to determine if a person's character has been unjustly targeted. This is because stereotypes are not based on the actual truth of the particular person they are attached to and stigmatises the individual. Their presence amidst character assassination is evidence of character injustice. Stereotype concepts will offer explainability as to why a certain word was tagged as being epistemically biased. 

We submit news article headlines to the CO-STAR and SBF models and receive an output of 6 potential stereotypes and 3 stereotype concepts the sentence is related to. Semantic similarity describes how closely related two items (e.g., 2 words or 2 sentences) are in terms of meaning. We use semantic similarity to determine how closely related each of the generated stereotype sentences in the list are to the original headline sentence. After obtaining the vector embedding for the sentences, we used a distance metric on the encoded vectors (cosine similarity metric from sentence transformers), to get the distance scores, then we ranked the outputs based on their semantic similarity to the headlines. The closest stereotype and stereotype concept to the sentence is output to the user as the potential stereotype and concept which casts doubt on the subject of the inputted text. 

\subsection{Lexicon Lookup}\label{ssec:methodlexicon} 
Once the tagger model has returned the top tagged word, we proceed to automatically look up and semantically search the tagged word in the epistemic lexicons from the social sciences - to discover the epistemic bias types it is associated with. These lexicons are from the collection of datasets we discuss in the dataset Section \ref{ssec:datasets}. When a tagged word is not found in the lexicons, we lemmatize said word (considering its context) to find its base word and find the stem word (removing the prefixes and suffixes). We then search for the lemmatized and stemmed words in the lexicons.

\subsection{Interactive Interface for Learning about Bias}\label{ssec:methodui} 

We propose the use of an interactive interface (UI) for editors of text content as a mechanism to keep human-in-the-loop control over outcomes of the models to mitigate any potential bias types and associated stereotypes, and make final editorial decisions prior to publishing their content.

\begin{figure}[H]
 \centering
 \includegraphics[scale=0.25]{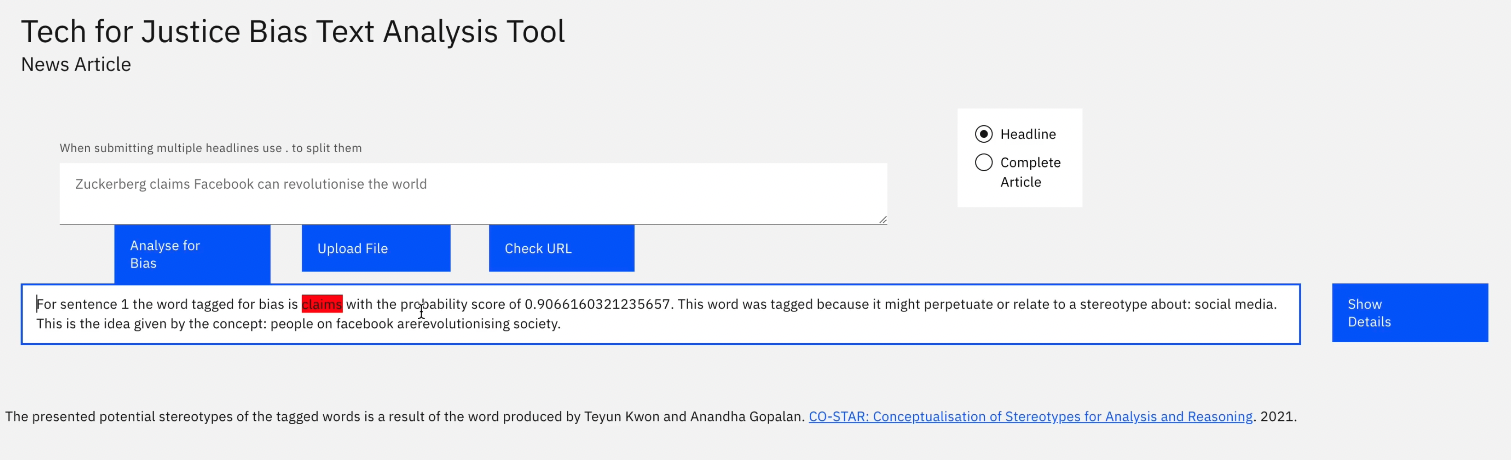}
 \caption{The interactive interface showing the highlighted tagged word and associated information for a sentence.}
 \label{fig:ui-1}
\end{figure}

During the analysis, we first leverage the nltk python library to remove any stop words (i.e., a, the, etc..). We also check the sentence length. We make an assumption here, that if the sentence is less than 3 words, then there is not enough context to analyse the text for potential stereotypes. However, if we have enough context, the content is sent to the epistemic tagger model to find the word with the highest probability of bias (see Figure \ref{fig:ui-1}). The conditions from the framework (Figure \ref{fig:framework}) are then implemented.

\begin{figure*}[h]
 \centering
 \includegraphics[scale=0.25]{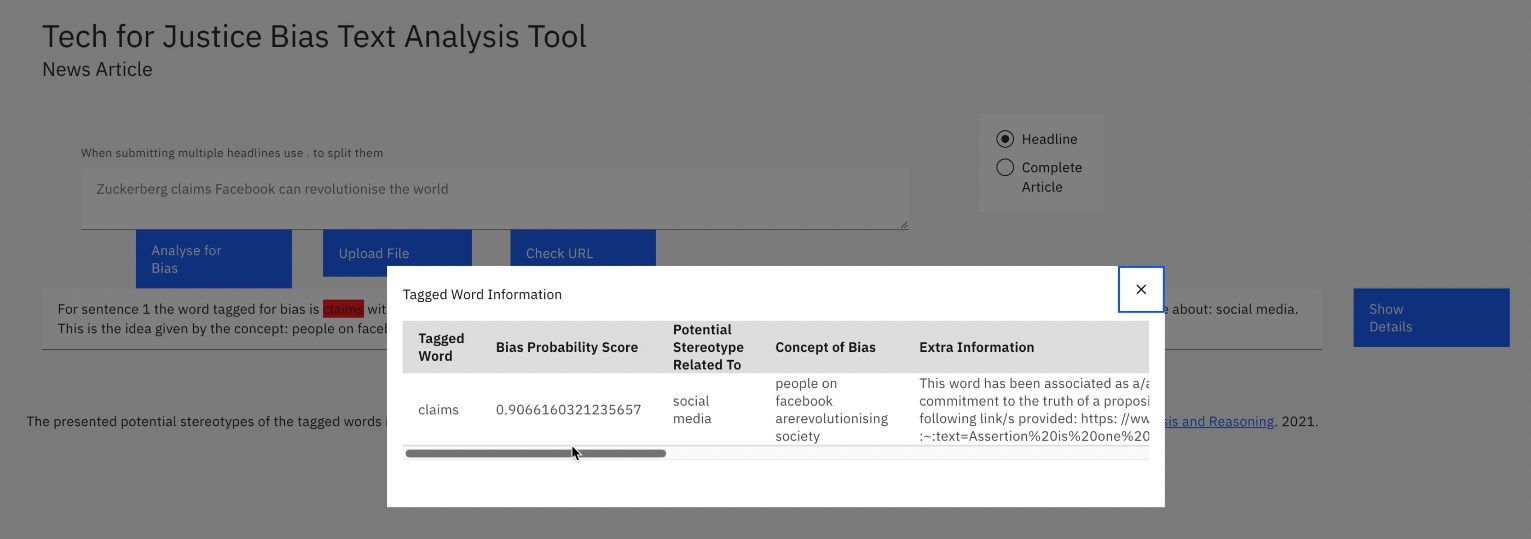}
 \caption{UI showing the tagged words, their associated probability for bias, the potential stereotype and stereotype concept and some explainable data created from the lexicon search.}
 \label{fig:ui-2}
\end{figure*}

In Figure \ref{fig:ui-1} and \ref{fig:ui-2} show screenshots of our UI after a user has submitted a text for analyzing. The word suspected of causing injustice is highlighted with the certainty score, an explanation of why the word was tagged. When the user clicks \emph{Show Details} they will see more details as displayed in Figure \ref{fig:ui-2}.

We display the types of associated epistemic bias detected to the user and provide links to resources which explains more on the specific types of biases as shown in Figure \ref{fig:ui-2}. See examples of these outputs in table \ref{tab:costar_output} and table \ref{tab:tagger_output} of Appendix \ref{sec:appendix}.

\subsection{Datasets} \label{ssec:datasets}

We leverage and use the bias data corpus\footnote{\url{http://nlp.stanford.edu/projects/bias/bias\_data.zip}} as created by \citet{pryzant2020automatically}. It contains a Wikipedia Neutrality corpus with Wikipedia articles that were annotated for neutrality by their editors, who adhere to NPOV guidelines \footnote{\url{https://en.wikipedia.org/wiki/Wikipedia:Neutral\_point\_of\_view}}. 

Lexicons used in the lexicon lookup were from social science research and collated in \citet{rpryzantgit}. We compiled these lexicons into one large dictionary and included some metadata about the lexicon i.e., source, creators, resources about the epistemic bias type, etc. Solely using an epistemic lexicon look up directly to find the biased words will be limited in its performance, for various reasons e.g., distributional shift, lexicons cannot scale well enough without requiring regular manual auditing, updating of the lexicon databases - when newer forms of subtle bias words arise, etc. There are various studies which show the limitations of using just a lexicon based approach alone, for example in \citet{lexvsuplearning}, the authors discuss how lexicon approaches perform less accurately than an end to end approach using BERT to detect gender based stereotypes. Therefore, it is beneficial to not solely lean on lexicons and we include it as a contributor in detecting injustices in our framework.

When it comes to Meghan Markle and Kate Middleton, two people associated to the British royal family by marriage, the media depicts several aspects of their lives which they share differently and often can be reflective of framing or character injustices\footnote{\label{buzzfeed}\url{https://www.buzzfeednews.com/article/ellievhall/meghan-markle-kate-middleton-double-standards-royal}}. We use actual headlines (some are shown in Table \ref{tab:headline_examples} with full list of examples in Table \ref{tab:headlines} of the Appendix \ref{sec:appendix}) from such depictions in our comparative test to further illustrate the detection and harms of these injustices. We scale the inputs to our framework from an initial 20 article headlines used in the comparative test by gathering 1645 articles from the Bing API (up to 100 articles for each topic from no more than 1 month back). We searched the following topics: ``Kate Middleton", ``Meghan Markle", ``Kate Middleton family", ``Meghan Markle family", ``Kate Middleton outfit", ``Meghan Markle outfit", ``Kate Middleton child", ``Meghan Markle child", ``Kate Middleton charity", ``Meghan Markle charity", ``Kate Middleton wedding", ``Meghan Markle wedding", ``Kate Middleton moving", ``Meghan Markle moving", ``Kate Middleton bullied", ``Meghan Markle bullied", ``Kate Middleton dress", ``Meghan Markle dress", ``Kate Middleton book", ``Meghan Markle book", ``Kate Middleton makeup", ``Meghan Markle makeup",``Kate Middleton health", ``Meghan Markle health", ``Kate Middleton photo", ``Meghan Markle photo", ``Kate Middleton money", ``Meghan Markle money". The results of these comparative tests are further discussed in our results (Section \ref{sec:results}).

\begin{table*}[h]
 \scriptsize
 \captionsetup{font=small}
 \setlength\defaultaddspace{0.66ex}
 \centering
 \begin{adjustbox}{min width=.7\textwidth}
 \begin{tabular}{p{0.06\linewidth} | p{0.6\linewidth} | p{0.08\linewidth} }
 \toprule
 Sentence No. & Headline & Subject \\
 \midrule
 1 & Meghan Markle spent a staggering £38,000 on her clothes for a charity trip & Meghan \\
 2 & Kate Middletons £100,000 Astonishing value of the dress that won a Prince's heart (and has hung in a wardrobe for eight years) & Kate \\
 3 & Meghan Markles beloved avocado linked to human rights abuse and drought, millennial shame & Meghan \\
 4 & Kates morning sickness cure? Prince William gifted with an avocado for pregnant Duchess & Kate \\
 \bottomrule
 \end{tabular}
 \end{adjustbox}
 \caption{Example headline sentences used. For a full list of the initial 20 examples, see Appendix \ref{appendix:full_comparative_test_table}}
 \label{tab:headline_examples}
\end{table*}


\subsection{Validation Survey} \label{ssec: valsurvey}
To validate our framework components, we conducted a validation survey with persons who should have some expertise in writing and/or editing. The survey included several questions aimed at evaluating the effectiveness of our framework components and pipeline to assist in automatically identifying areas for improvement for journalists and editors striving for fairness and neutrality in their text. Therefore we conduct a survey to validate the credibility and viability of our approach. We sought participants who work/volunteer in a writing/editing centers, are Senior/Grad Level journalism students, education students, educators, journalists, editors, publishers, bloggers, authors/ editors/reviews of a published work, or have operated as a reviewer for a journal/conference that accepts publications. The survey was approved by our local Institutional Review Board (IRB) and conducted through an online survey form. The respondents had unlimited time to respond and could only submit a single response. We had a total of 30 respondents, of which 8 were removed due to completing multiple surveys, improperly answering the questions (i.e., giving random answers to the survey questions), or refusal to adhere to directions of the investigator. The survey presented 16 of the headlines from Table \ref{tab:headlines} (8 pertaining to Kate, 8 pertaining to Meghan) and asked these questions about the headlines: 
\begin{enumerate}
    \item Main epistemic biased word is:, What alternative word or phrase will you use?
    \item If there is an injustice present, how would you label it? (Framing Injustice, Character Injustice, Testimonial Injustice, None)
    \item Is there sarcasm present in this headline? (yes, no)
    \item What is the sentiment of this headline? (Positive, Neutral, Negative, Don't know)
    \item Which stereotype presented here is the most related to this headline?
\end{enumerate}

Questions 1 and 2 were freeform, short-answer, while questions 3 - 6 were multiple-choice responses. Question 6 asked for the stereotypes based on the top 4 responses from the SBF and CO-Star Models. Lastly, to validate our assumption on adding a TMI feature, we tested these prompts which are from our TMI headline dataset and the modified versions of the headlines (containing less descriptors for the first headline and more descriptors for the remaining headlines): \textbf{ ``Brazil: evangelical superstar expelled from Congress over alleged role in husband’s murder", ``Woman unseated from Congress due to alleged role in husband’s murder", ``Last Man Standing: Suge Knight and the Murders of Biggie and Tupac review - a rueful return to Death Row", `Suge Knight and the Murders of Biggie and Tupac, former opponents and rappers, review - a shameful return to Death Row", ``Grace Millane trial concludes with New Zealand male convicted as guilty of murdering the British woman backpacker", and ``Grace Millane trial: New Zealand man found guilty of murdering British backpacker"} - with the following questions: 
 
 \begin{enumerate}
     \item What is the sentiment of this headline? (Positive, Neutral, Negative, Don't know)
     \item How do you perceive the subject of the text?
     \item What word is framing the sentence here?
     \item How difficult is it for you to find the framing word (1-5, 5 being most difficult)? (1,2,3,4,5)
\end{enumerate}

Questions 2 and 3 were freeform, short-answer; while questions 1 and 4 were multiple-choice responses.

\section{Results and Discussions}\label{sec:results}

\subsection{Tagger Model Ablation Study}\label{sec:abalationstudy}
We performed an ablation to analyse the fine-tuned tagger model performance when there is no expert feature as proposed by \citet{recasens2013linguistic} and when they are included. Part of the training experiment was also to determine if there are any benefits from including the TMI feature into the input data. We carry out the training on 23,000 training samples from the bias-data training set and 700 validation and 1,000 test samples. Training was done using 4 CPUs and 1 GPU. We used a learning rate of 3e-5 and initially trained for 10 epochs. 

\begin{table}[h]
 \scriptsize
 \captionsetup{font=small}
 \setlength\defaultaddspace{0.66ex}
 \centering
 \begin{adjustbox}{min width=0.4\textwidth}
 \begin{tabular}{p{0.1\linewidth} | c | c }
 \toprule
 \hline
 \multicolumn{3}{c}{Tagger Model Ablation Experiments} \\
 \hline
 Kind & Evaluation Accuracy(\%) & Evaluation Loss \\
 \hline
 \multirow{3}{4em}{basic} & 72.54 & 0.0758 \\ & 72.39 & 0.0766 \\ & 73.13 & 0.0851 \\ 
 \hline
 \multirow{3}{4em}{+ expert features} & 75.04 & 0.0745 \\ & \textbf{74.16} & \textbf{0.0734} \\ & 72.83 & 0.0867 \\ 
 \hline
 \multirow{3}{4em}{+ tmi} & 74.79 & 0.0730 \\ & \textbf{*74.63} & \textbf{*0.0703} \\ 
 & 74.63 & 0.0822 \\ 
 \hline
 \bottomrule
 \end{tabular}
 \end{adjustbox}
 \caption{Ablation study results showing values for the first 3 epochs. Note: A high score is better for \textit{accuracy} and a lower score is better for \textit{loss}. Bold values indicate when the evaluation loss was less than the previous step. }
 \label{tab:ablation-results}
\end{table}

\begin{figure*}[h]
 \centering
 \includegraphics[scale=0.3]{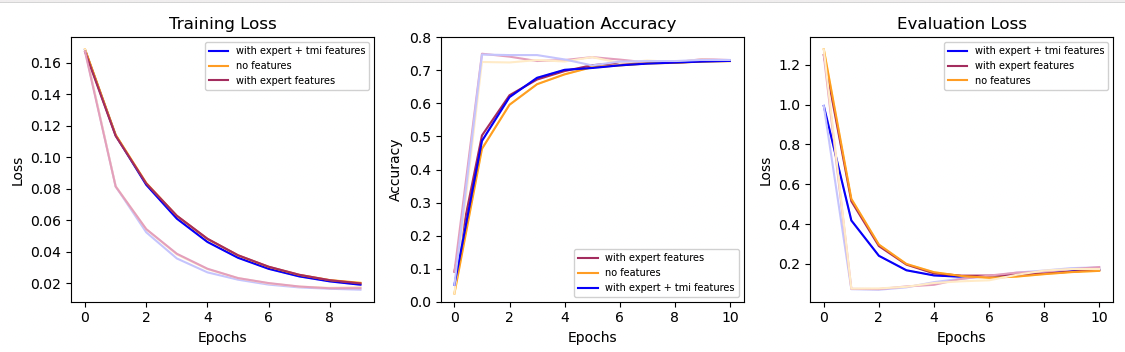}
 \caption{Training and Validation Results for the Tagger Model using 10 epochs. \textit{\textbf{no features}} = the expert features as mentioned in Section \ref{ssec:methodtagger} was not added, \textit{\textbf{with expert features}} = combine the expert features as mentioned in Section \ref{ssec:methodtagger} and \textit{\textbf{with expert + tmi features}} = combine the expert features whilst using the data samples containing TMI class \textit{no tmi} or \textit{tmi}}
 \label{fig:tagger_result_10}
\end{figure*}

As shown in Table \ref{tab:ablation-results} and Figure \ref{fig:tagger_result_10}, we observe a lower training loss value between epoch 3-6 during the initial training using the dataset than when we included the TMI feature. TMI also gives a better evaluation accuracy and evaluation loss as compared to the others before any overfitting is observed (which causes the evaluation loss to exponentially increase from about epoch 3). This overfitting occurs as a result of using too many epochs during the initial training of the tagger model. When fine-tuning a BERT model, a good practice as suggested by the authors of BERT is to use between 2 and 4 epochs. In light of this, and from the observations made from the initial training experiments, we save and use as the best point the tagger model after epoch 2.

\subsection{Comparative Test} \label{ssection: compare}
We performed a comparative test to illustrate how our framework shows injustices e.g., character and framing injustice. Particularly, we capture the type of epistemic bias attached to the tagged word in each entry, observe if a relevant stereotype is associated with the inputs, and observe the depiction of different subjects. In this study, we look at Meghan Markle and Kate Middleton as our subjects of interest. Meghan is associated to the British royal family, by marriage, and resigned from her royal duties due to abuse from the media and royal family (e.g., the Firm). Kate is also a member of the British royal family, by marriage, and was promoted to Princess of Wales. Even before Meghan Markle had resigned from her royal duties, the media sought to minimize her experience and diminish her comments and character.  We often see instances of Meghan as the subject of an article and the media using sarcasm and criticism towards her, thus character injustice. Such acts have led readers to having tainted images of Meghan and even not believing her statements or actions, thus testimonial injustices. Yet, for the same and similar topics of concern, media members speak charmingly about Kate, thus leading to framing injustices shown. It is important to note here, that Kate Middleton and Meghan Markle shared similar positions, interests, and abilities. With this, we see them as good subjects to observe in our comparative test.

\begin{table*}[hbt!]
 \scriptsize
 \centering
 \begin{tabular}{p{0.03\linewidth} | p{0.055\linewidth} | p{0.28\linewidth} | p{0.075\linewidth} | p{0.3\linewidth} | p{0.08\linewidth} }
 \toprule
 No. & Subject & Stereotype(S) & Distance(S) & Stereotype Concept(SC) &  Distance(CS) \\
 \midrule
 1  & Meghan & personal spending habits & 0.3457 & women should spend money on clothes & 0.4914 \\
 2  & Kate & women should be dressed like brides & 0.3259 & women are property & 0.2278 \\
 3  & Meghan & feminism & 0.3663 & sexism & 0.2523 \\
 4  & Kate & british women are marginalized for a joke & 0.2083 & pregnancy & 0.3918 \\
 \bottomrule
 \end{tabular}
 \captionsetup{justification=centering}
 \caption{Table showing the top ranked potential stereotype and stereotype concept (semantic distance \(> 0.3\) to the headline. The \texttt{No.} column represents the headline number from Table \ref{tab:headline_examples}. See full list of result in Table \ref{tab:costar_output} in Appendix \ref{appendix:full_comparative_test_table}. }
 \label{tab:stereo_distance}
\end{table*}

\begin{figure*}[h]
 \centering
 \includegraphics[scale=0.35]{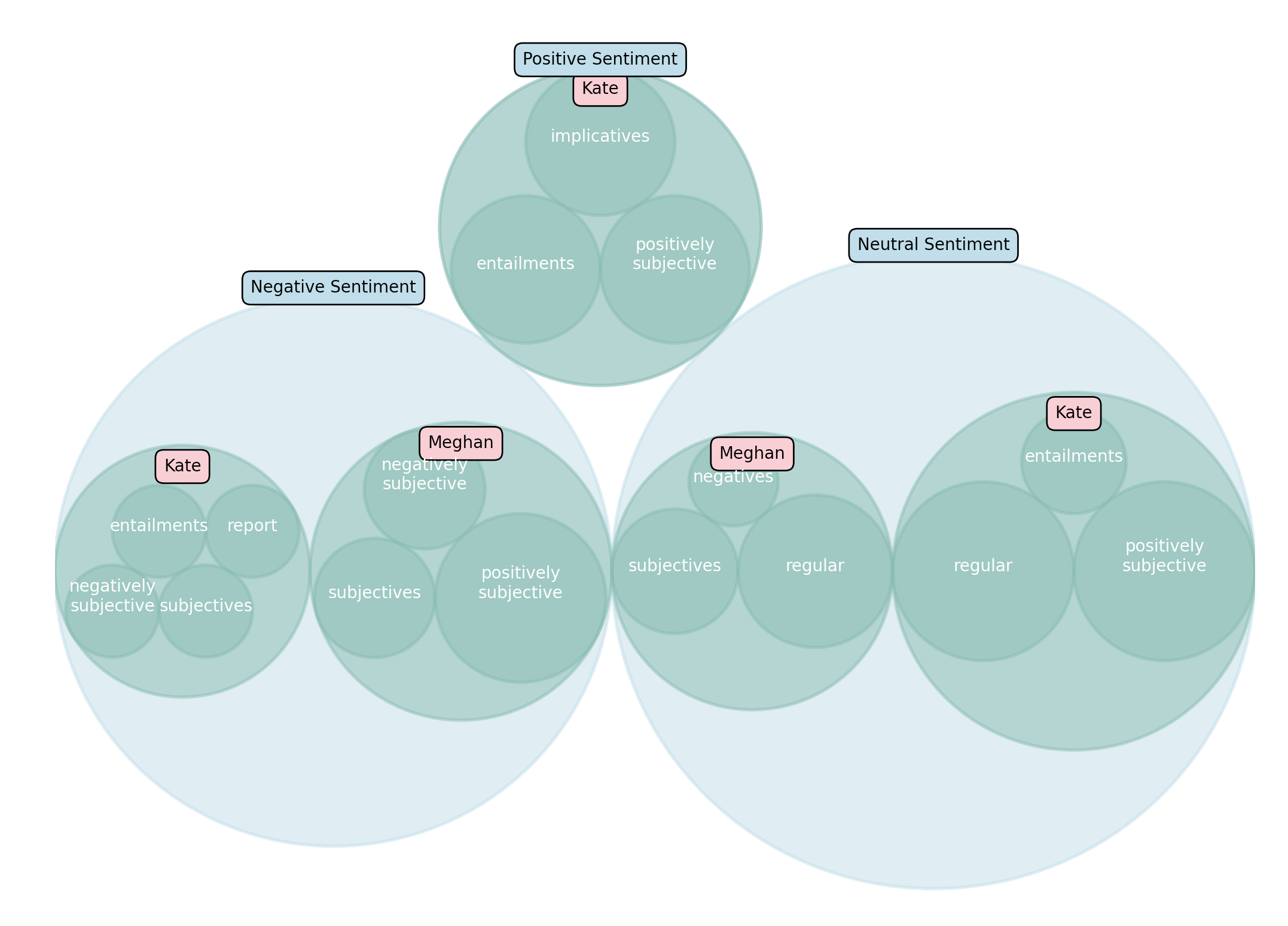}
 \caption{Epistemic Bias Types - Comparative Test}
 \label{fig:comptestcircles}
\end{figure*}

In Figure~\ref{fig:comptestcircles}, we plotted results of an experiment where we initially analysed 10 article headlines of Meghan Markle and 10 articles of Kate Middleton through our framework. Each article was chosen because it discusses a topic which is common between each of the two subjects and allows us to see the differences in how the two subjects are spoken about concerning that topic. It has been acknowledged by several outlets that these particular articles which we have mentioned are unjust\textsuperscript{\ref{buzzfeed}}. Majority of these articles are seen in the aforementioned articles to show these comparisons, but also a few additional articles were chosen by our team. In this plot, we capture the subject of the headline (Meghan/Kate), sentiment of the entire headline sentence, and the associated epistemic bias types for each headline as shown in table \ref{tab:tagger_output} located in Appendix \ref{sec:appendix}. See the complete list of headlines in Table \ref{tab:headlines} in Appendix \ref{sec:appendix}. The light blue circles represent the overall sentiment of the headlines, we will refer to them as the sentiment headline level. We use the 3 categories of positive, neutral, and negative sentiments. Within each sentiment headline level, we capture the subject of the headline, we will call this the subject level. Note that in Figure \ref{fig:comptestcircles}, Kate takes up the entire positive sentiment space, there were no examples that were detected to have positive sentiments for Meghan. Within each subject level, we capture the epistemic bias types associated with the tagged words in the headlines for each subject; we will refer to this as the bias-type level. This illustration shows us how much space is filled by each subject.

From this plot, we can see the headline sentiment level, the negative sentiment circle is mostly filled with articles about Meghan and the most common bias type is \emph{positively subjective}. The intuition here is that many articles about Meghan contain sarcasm, though this is not the only indicator. There was one article about Kate which had negative sentiment and each of the bias types in her circle were identified in the headline, this is why they are all equal in size. In looking at the headline sentiment level, the positive sentiment circle is completely filled with articles about Kate and the most common bias type is \emph{implicatives}. The intuition here is that, in articles which have common topics between Meghan and Kate, only Kate is talked about in a suggestively positive way. In looking at the headline sentiment level, the neutral sentiment circle is mostly filled with articles about Kate and the most common bias type is \emph{positively subjective} and \emph{regular}. Recall, regular is the epistemic bias type assigned to detected words which do not appear, nor do similar words appear in the lexicon lookup. We can also observe, the articles which appear for Meghan with neutral sentiment are \emph{negatively subjective} in nature, informing us they likely are not sarcastic but have negative suggestions. For Kate, she is talked about in a \emph{suggestively positive} way, even in the neutral sentiments. This analysis confirms that when there is a topic that Meghan and Kate share - there was a use of words for Meghan which are sarcastic and cause framing and character injustices to her as a subject. Both of which can contribute to someone taking Meghan less seriously, thus causing testimonial injustice.

We manually evaluated the relevance of the generated stereotypes to the headlines as shown in table \ref{tab:stereo_distance} (see appendix table \ref{tab:costar_output} for the full table). A lower semantic similarity scores implies the generated stereotypes are irrelevant to the headlines. Where the semantic similarity distance was greater than 0.3, we observe that 5 out of 7 of such entries related to Meghan as the subject and only 2 related to Kate.  More interestingly, we observed that one of the potential stereotypes relating to Meghan (\textit{personal spending habits}) aimed at a personal attribute. Which can be seen as an indication of character assassination, thus character injustice.

\subsubsection{\textbf{Scaling the comparative test}}\label{sssection:scale_compare}

After comparing the few headlines for both Kate and Meghan in \ref{ssection: compare} above, we used the process of getting more data into the framework as shown in Figure \ref{fig:crawling} to extract about 1600 new headlines for both subjects into the framework. Our aim of doing this is to show that the framework can be scaled to help aide the automatic analysis for potential subtle biases and injustice in text. 

\begin{figure} [H]
 \centering
 \begin{subfigure}[b]{1\textwidth}
   \centering
  \includegraphics[width = 0.9\textwidth]{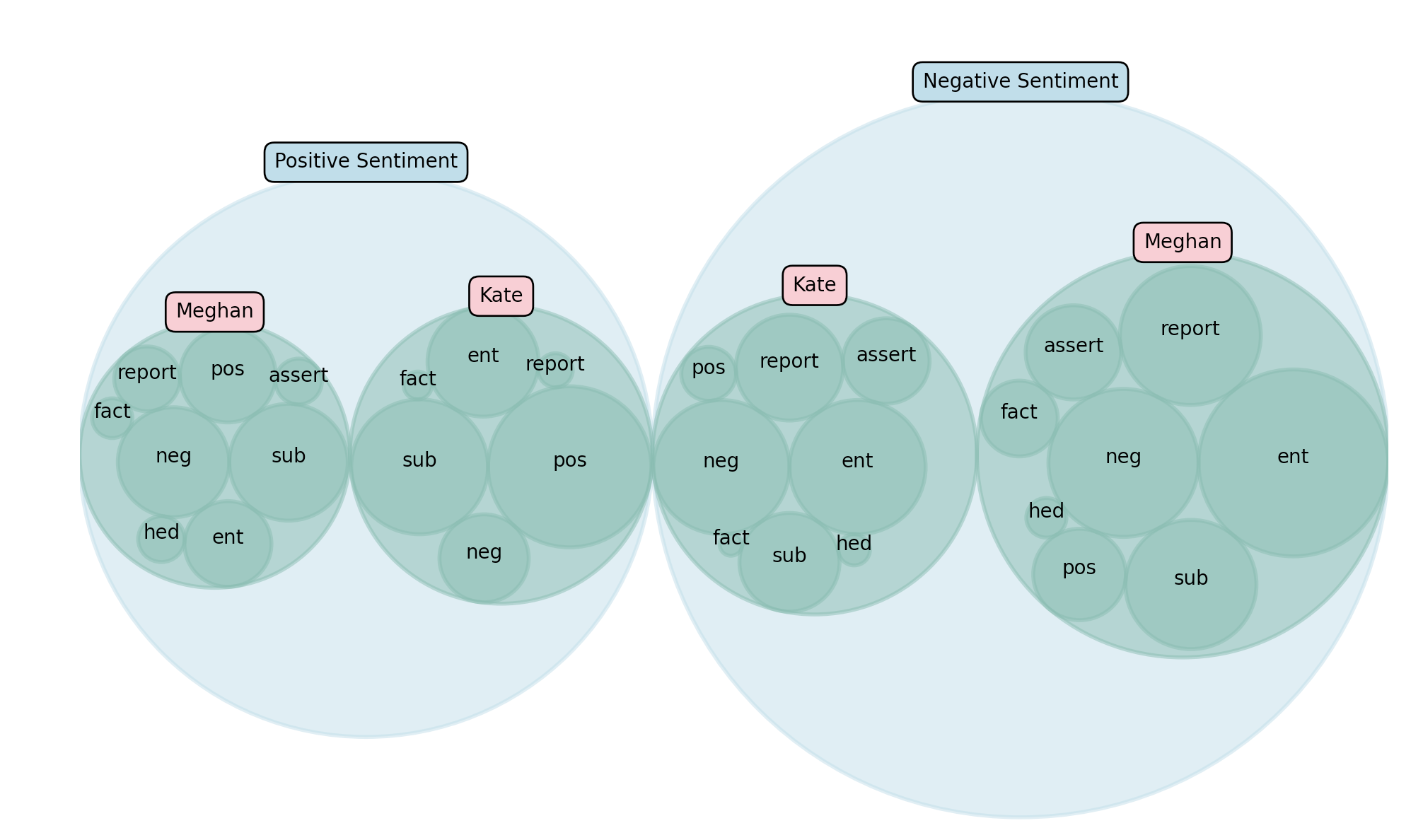}
   \caption{Positive and Negative Sentimental Context - Scaled Comparative Test. Note the epistemic bias types are abbreviated as such for spacing concerns - Assert: Assertives, Ent: Entailments, Hed: Hedges, Fact: Factives, Neg: Negative, Pos: Positive, Sub: Subjective.}
  \label{fig:pos_neg_large_comptestcircles}
 \end{subfigure}
 \medskip
 \begin{subfigure}[b]{1\textwidth}
   \centering
  \includegraphics[width = 0.7\textwidth]{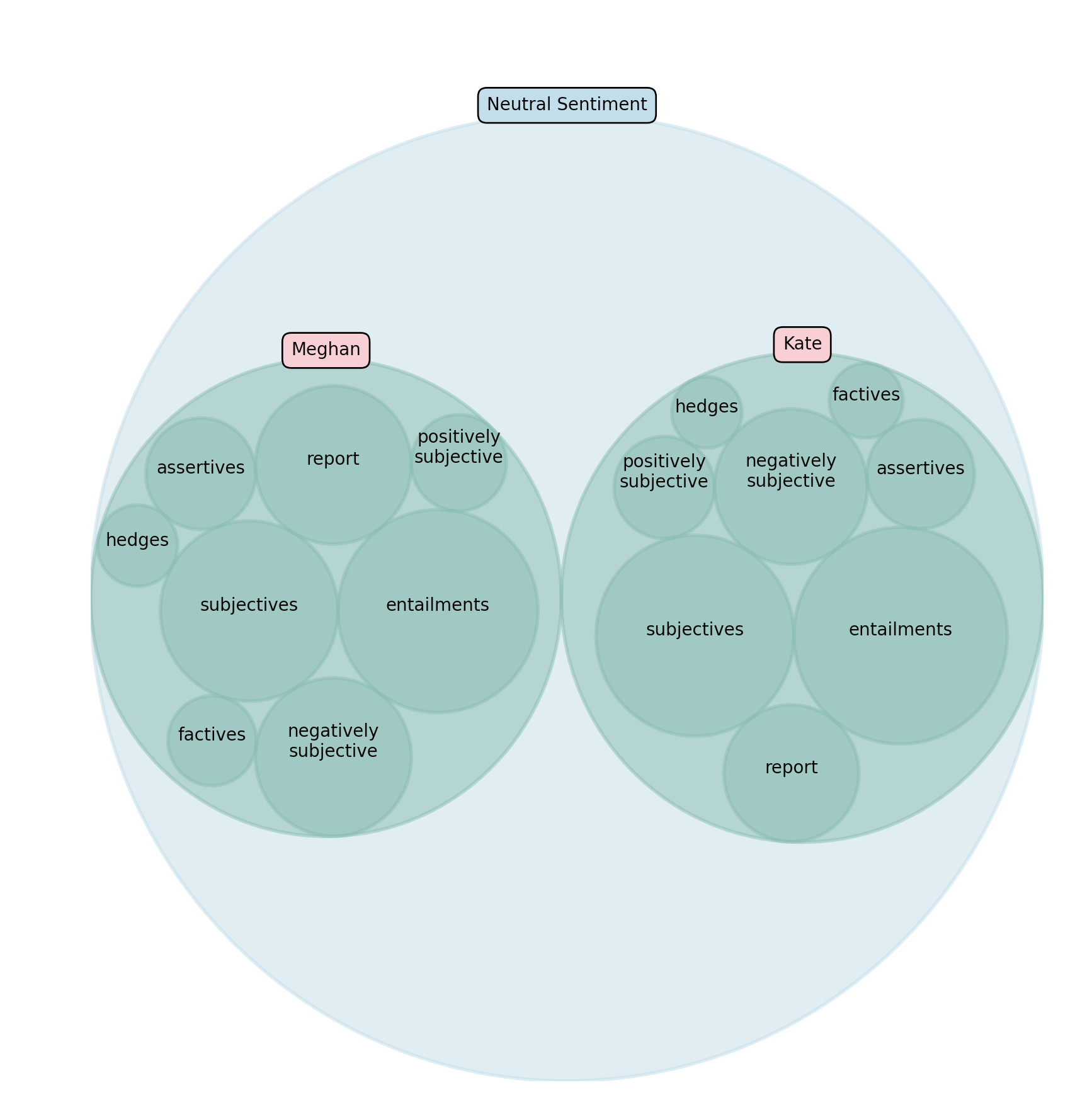}
  \caption{Neutral Sentimental Context - Scaled Comparative Test}
  \label{fig:neu_large_comptestcircles}
 \end{subfigure}
 \caption{Epistemic Bias Types - Scaled Data Comparative Test.}
 \label{fig:large_comptestcircles}
\end{figure}

In Figures \ref{fig:pos_neg_large_comptestcircles} and \ref{fig:neu_large_comptestcircles}, we see the plots obtained from scaling this analysis. Supporting our previous discussion, the observation from the \texttt{Negative Sentiment} circle is that Meghan often has more implicit subjective biases in a negative sentimental context compared to Kate. Likewise, for Kate we observed that the implicit subjective biases detected in the text associated with her are mainly used in a positively sentimental context (she takes up more space in the overall \texttt{Positive Sentiment} circle compared to Meghan). Most of the sample headline sentences had a \texttt{Neutral Sentiment}, hence the reason why the outer circle in Figure \ref{fig:neu_large_comptestcircles} showing the neutral sentimental context is extremely large. We observed that both subjects had almost equal amount of the epistemic bias types in this neutral context. 

\subsection{Human Survey Discussions} \label{ssec: valsurveyres}
\begin{figure*}[h]
 \centering
 \includegraphics[scale=0.37]{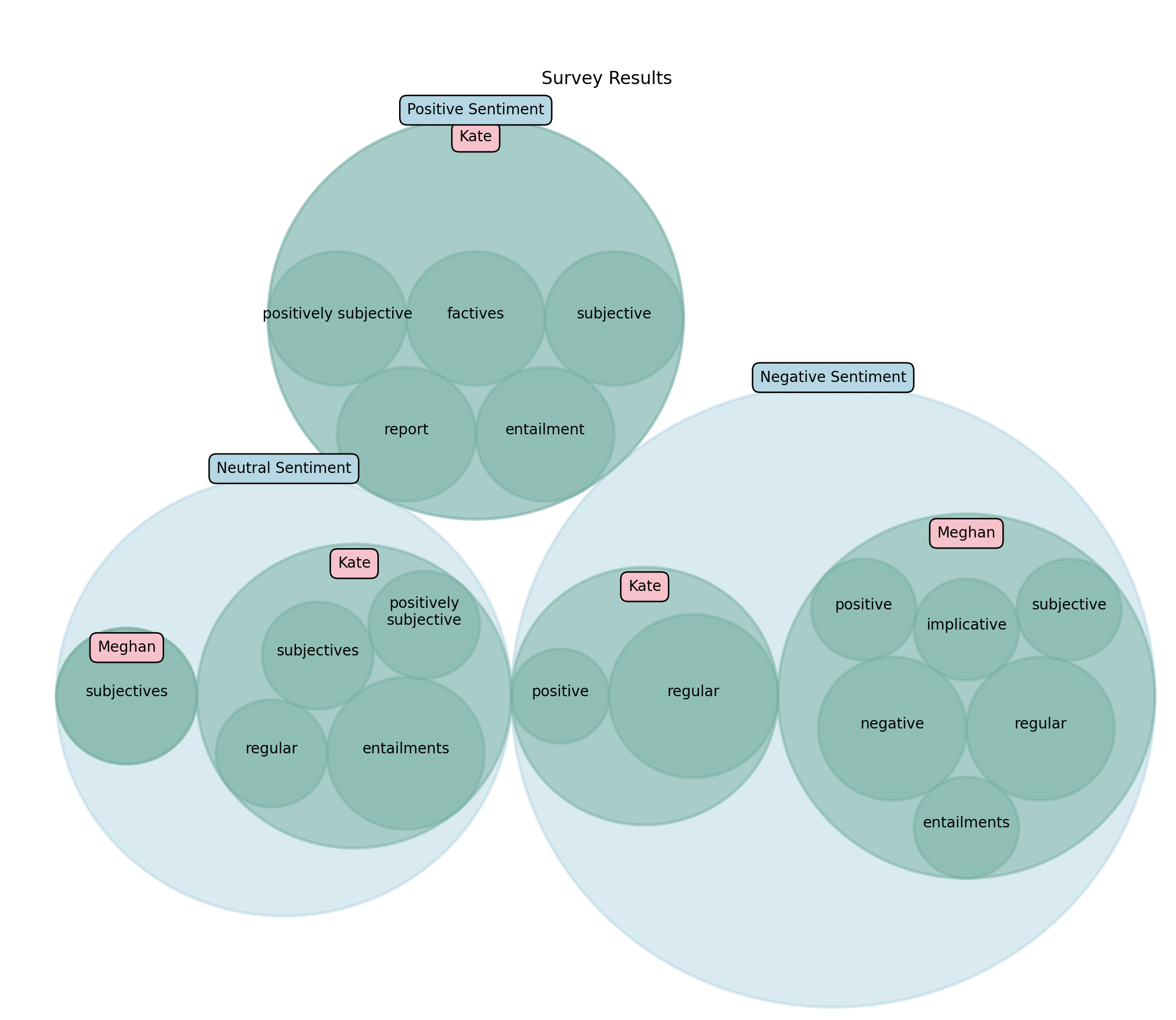}
 \caption{Comparative Test for Survey Participant Defined Tagged Words and Headline Sentiments}
 \label{fig:surveycircles}
\end{figure*}

Our 22 valid respondents were made up of 1 person who had operated as a reviewer for journal, 5 educators, 3 bloggers, 3 editors/publishers, 1 who had operated as a reviewer for a conference publication, 6 journalists, 1 author of a published work, and 2 PhD Students in Journalism/Education/English. We find the tagger model of the framework agrees with the biased word choice of humans 52\% of the time, the stereotype 41.2\% of the time (though some choices differed, still related stereotypes in 52.9\% of the time), and the sentiment 66.67\% of the time. So, majority of the time, the components of the framework agreed with human analysis. In Table \ref{tab:surveysumcomparisonSHORT}, there is a short list of the top survey answers to the questions and the output of the framework (see full list in Appendix \ref{sec:appendix}) Figure \ref{tab:surveysumcomparison}. When comparing the trend of the headlines from Figure \ref{fig:comptestcircles} and Figure \ref{fig:large_comptestcircles}, we see from the survey results in Figure \ref{fig:surveycircles} similarly that respondents were able to detect similar proportions of sentiments for each subject of the text. We do see people tended to decide more definitively that headlines were either positive or negative rather than neutral, causing those proportion of headlines to be greater. We see more diversity in the types of epistemic biases present in the neutral and positive headlines for Kate and more diversity in the types of epistemic biases present in the negative headlines for Meghan. The participants also determined that for 3 of the headlines concerning Meghan sarcasm was equally suspected and not suspected, for 2 of the 5 remaining articles about Meghan sarcasm was detected, and for 3 of the headlines concerning Kate sarcasm was detected. This might further solidify that these instances of injustice are subtle and can be difficult to detect. The main feedback from participants of the survey was it took them longer than they would like to analyze, digest, and decide on each element encompassed in the framework. Several participants indicated they had to take breaks and return to the survey while one valid non-participant indicated they wanted to participate in the survey but that it took them too much time to answer the questions, resulting in them recusing themselves from the survey. This shows us there is a need for our novel automated framework which gives similar results to the human analysis, majority of the time. Another observation from the study is that when answering Question 2 (\emph{What alternative word or phrase will you use?}), many of the respondents used words which could also perpetuate some type of injustice while others simply thought it best to remove the word causing injustice all together. Two respondents who identified as a journalist and someone who had operated as a reviewer for journal, noted that though some words caused injustices, they did not want to suggest an alternative word for any headline. Another respondent who identified as an educator similarly did not give alternative word choices for several of the scenarios but found injustice. This could be due to lack of confidence for finding a better word choice or being indifferent to the injustice occurring.  

When reviewing the participant's answers for the TMI portion of the survey, participants indicated it easier for them to find the word framing the sentence and causing injustices when there were more descriptive words present. This concurs with our hypothesis mentioned in Section \ref{sec:abalationstudy}. They also tended to use less harsh words when describing their thoughts of the subjects of the headlines with less descriptors, even when they felt it was more neutral and less negative.

\begin{table}[H]
\fontsize{13pt}{13pt}\selectfont
\centering
\resizebox{0.9\columnwidth}{!}{
\begin{tabular}{|c|l|l|}
\hline
\textbf{Headline}                                  & \multicolumn{1}{c|}{\textbf{Participant’s choices (\# of participants)}}                    & \multicolumn{1}{c|}{\textbf{Framework Component’s choice (confidence)}}             \\ \hline
\textit{\begin{tabular}[c]{@{}c@{}}Meghan Markle spent a staggering £38,000 on her clothes \\  for a charity trip\end{tabular}}                  & \begin{tabular}[c]{@{}l@{}}Tagged word: staggering (20)\\ epibias\_type: subjective\\ sentiment: negative (15)\\ stereotype: personal spending habits (18)\\ Injustice Present: Character (14)\\ Alternative word: significant (3)\end{tabular}                                & \begin{tabular}[c]{@{}l@{}}Tagged word: staggering (0.999498) \\      epibias\_type: subjective\\      sentiment: negative\\      stereotype: personal spending habits\end{tabular}                       \\ \hline

\textit{\begin{tabular}[c]{@{}c@{}}Kate Middleton's £100,000 Astonishing value of the dress that \\  won a Prince’s heart (and has hung in a wardrobe for eight years)\end{tabular}}                                                              & \begin{tabular}[c]{@{}l@{}}Tagged word: astonishing (20)\\  epibias\_type: subjective\\  sentiment: negative (15)\\  stereotype: personal spending habits (18)\\  Injustice Present: Character (14)\\  Alternative word: significant (3)\end{tabular}                                               & \begin{tabular}[c]{@{}l@{}}Tagged word: astonishing (0.999342) \\      epibias\_type: positively subjective\\      sentiment: neutral\\      stereotype: women should be dressed like brides\end{tabular}  \\ \hline

\textit{\begin{tabular}[c]{@{}c@{}}Meghan Markle's beloved avocado linked to human rights abuse \\  and drought, millennial shame\end{tabular}}                                                                                                   & \begin{tabular}[c]{@{}l@{}}Tagged word: beloved (18)\\  epibias\_type: positively subjective\\  sentiment: negative (19)\\  stereotype: feminist are hypocrites (11)\\  Injustice Present: Character (13)\\  Alternative word: " " (2)\end{tabular}                                                 & \begin{tabular}[c]{@{}l@{}}Tagged word: beloved (0.997946) \\      epibias\_type: positively subjective\\      sentiment: negative\\      stereotype: feminism\end{tabular}                                \\ \hline

\textit{\begin{tabular}[c]{@{}c@{}}Kate’s morning sickness cure? Prince William gifted with \\  an avocado for pregnant Duchess\end{tabular}}                                                                                                     & \begin{tabular}[c]{@{}l@{}}Tagged word: cure (8), gifted (8)\\  epibias\_type: entailments, positively subjective\\  sentiment: neutral (10)\\  stereotype: British women are marginalized for a joke (11)\\  Injustice Present: Character (8)\\  Alternative word: given (4)\end{tabular}          & \begin{tabular}[c]{@{}l@{}}Tagged word: gifted (0.877285) \\      epibias\_type: positively subjective\\      sentiment: neutral\\      stereotype: British women are marginalized for a joke\end{tabular} \\ \hline

\end{tabular} 
}
\caption{Short List of Comparing Survey Results for Participants and Framework Outputs} \label{tab:surveysumcomparisonSHORT}
\end{table} 

\section{Conclusions}\label{sec:conclusions}
We describe a novel framework which uses a unique combination of different NLP techniques to detect character, testimonial, and framing injustices in text. These forms of injustices are often subtle and hard to quickly detect. The framework includes a fine-tuned BERT based epistemic tagging model, a GPT based stereotype generative model and SBF model, semantic similarity searching and lexicon lookup of epistemic biased words from the social science field. We provide qualitative empirical evidence as a justification for using this framework to detect such injustices in text. Further, we show our proposed UI (aimed at helping editors and authors of text content), by automatically detecting and explaining potential bias types and injustices that might be present in their content. We anticipate this UI will encourage them to take necessary, preventative steps in avoiding unjust acts pre-publication. 

 



 

\section*{Limitations}
\begin{itemize}
 \item Learning Better Patterns for injustices:- 
 Identifying a single word which potentially cause bias in a given sentence is only a start. We acknowledge that multiple words or phrases in a given sentence might be the culprit and the ability to tag multiple words or phrases will take this work further.
 \item Veridicality Assessment:- We intend to incorporate veridicality assessment in our framework to assess if a statement is actually factual, which will allow for a more accurate analysis.
 \item Evaluating Generative Stereotype Models:- We needed to manually evaluate the relevance of the generated stereotypes to our headlines. In the future, we would like to have evaluation metrics to help with analyzing the stereotypes generated by the CO-Star and SBF models.
\end{itemize} 

\section*{Ethics Statement} \label{sec:issues}
A critical area for bias in systems and models' design often stem from a given human's intrinsic biases. They are usually a reflection of ourselves. One possible solution to this problem is to ensure that a diverse group of individuals are involved from the inception of the solutions design to the testing phase of such technical solutions---this is one reason why we gathered a diverse set of authors to be involved in the discussions presented in this paper. Further, we were intentional in sending our survey to venues who serve marginalized communities (e.g., writing centers at Latino serving institutions, educational institutions who have a majority of minoritized student body populations, Black graduate student associations, Asian journalist associations, etc...).

A second area for bias surrounds the definition of terms and assumptions biases. A way we attempt to resolve this is by considering the existing consensus definitions of epistemic bias from the social sciences, where media bias has been studied for decades. 

A third potential area for bias can occur in the training and validation datasets, as well as in the models used for implementing the proposed solution. For example, the sourced datasets do not contain enough semantic information that is very reflective of the injustices considered. One possible way to combat this is to investigate ways to build more robust training datasets or leverage models that do not require lots of training examples to generalise properly.

Broader problems may arise in any situation where technology is naively applied to solve a societal issue. As envisaged, our framework should be applied as a means to help people working in the media improve their output with respect to bias and injustices. However, as warned by Goodhart's law~\citep{manheim2018categorizing}, if the measures and metrics suggested here become targets, they will cease to be useful. For example, in situations where experts deliberately bias their content the tool can become beneficial to the readers instead, so that they are aware of potential biases when reading an article. However, a main purpose of our proposed concept and tool is to help journalists who are aware that they might use biased terms but do so unintentionally. On the other hand, we cannot control the adoption of our tool. It will however help the editors who are checking for bias using manual means to quickly detect such biases. For example, the authors or editors of Wikipedia contents, have to follow and adhere to a NPOV (neutrality point of view) encyclopedia when writing or editing article contents, to ensure a neutral view point, and such a tool will be beneficial to them.



\bibliographystyle{ACM-Reference-Format}
\bibliography{software}

\onecolumn
\appendix
\section{Appendix}
\label{sec:appendix}



\begin{figure}[H]
 \centering
 \includegraphics[scale=0.4]{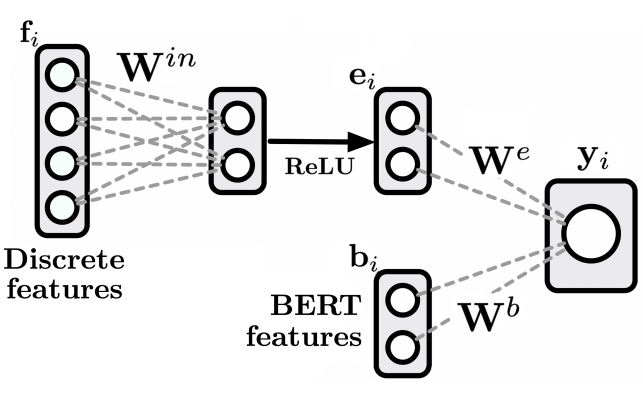}
 \caption{The tagger model computes logits \(y_i\) using discrete feature \(f_i\) and BERT embedding \(b_i\)}
 \label{fig:tagger-detection-model}
\end{figure}

In Figure \ref{fig:tagger-detection-model}, we have a diagram showing how the tagger model computes logits \(y_i\) using discrete feature \(f_i\) and BERT embedding \(b_i\) \cite{pryzant2020automatically}. 

\vspace{-3mm}

In Table \ref{tab:headlines}, we have all the headlines used in the comparative study between Meghan Markle and Kate Middleton. Several news outlets have also noted that these articles are used unjustly to depict one person differently than the other on a shared topic, playing on stereotypes about the person.

\begin{table*}[htbp]

 \captionsetup{font=small}
 \setlength\defaultaddspace{0.66ex}
 \centering
 \begin{adjustbox}{max width=1\textwidth}
 \begin{tabular}{ p{0.08\linewidth} | p{0.8\linewidth} | l }
 \toprule
 Sentence No. & Headline & Subject \\
 \midrule
 1 & Meghan Markle spent a staggering £38,000 on her clothes for a charity trip & Meghan \\
 2 & Kate Middletons £100,000 Astonishing value of the dress that won a Prince's heart (and has hung in a wardrobe for eight years) & Kate \\
 3 & Meghan Markles beloved avocado linked to human rights abuse and drought, millennial shame & Meghan \\
 4 & Kates morning sickness cure? Prince William gifted with an avocado for pregnant Duchess & Kate \\
 5 & Kate and William 'packed up the kids' in search of 'privacy' at new Windsor estate & Kate \\
 6 & Why Prince Harry and Meghan Markle's Potential Plan to Protect Their Family is Incredibly Unfair to Archie & Meghan \\
 7 & Prince Harry and Duchess Meghan Think Moving to Canada Will Give Archie a Normal Upbringing & Meghan \\
 8 & Not long to go! Pregnant Kate tenderly cradles her baby bump while wrapping up her royal duties ahead of maternity leave - and William confirms she's due 'any minute now' & Kate \\
 9 & Why can't Meghan Markle keep her hands off her bump? Experts tackle the question that has got the nation talking: Is it pride, vanity, acting - or a new age bonding technique? & Meghan \\
 10 & Kate Middleton Wore a Bardot Dress to the 'Top Gun' Premiere & Kate \\
 11 & Kate Middleton's homegrown bouquet of lily of the valley follows royal code & Kate \\
 12 & Royal wedding: How Meghan Markles flowers may have put Princess Charlottes life at risk & Meghan \\
 13 & Duchess Kate reveals her favourite photo of son Prince Louis & Kate \\
 14 & How Meghan and Harry Ripped Up Royal Tradition on Birthday Photos of Archie & Meghan \\
 15 & Kate Middleton Debuts New Sapphire Earrings That Belonged to Princess Diana and Debunks a Rumor! & Kate \\
 16 & Meghan Markle Just Casually Rewore Her \$16,500 Royal Wedding Earrings in NYC & Meghan \\
 17 & Kate and Wills Inc: Duke and Duchess secretly set up companies to protect their brand - just like the Beckhams & Kate \\
 18 & A right royal cash in! How Prince Harry and Meghan Markle trademarked over 100 items from hoodies to socks SIX MONTHS before split with monarchy - with new empire worth up to £400m & Meghan \\
 19 & Kate Middleton deviated from royal 'cleavage protocol' with daring neckline & Kate \\
 20 & Meghan Markle Just Wore a Plunging Gown With a Thigh-High Slit on the Red Carpet & Meghan \\
 \bottomrule
 \end{tabular}
 \end{adjustbox}
 \caption{ Comparative Headline Sentences Used}
 \label{tab:headlines}
\end{table*}

\label{appendix:full_comparative_test_table}
Table \ref{tab:tagger_output} in this section shows the full output of tagging each headline sentence used in the comparative test. The \texttt{No.} column corresponds to the sentence number, \texttt{Subject} refers to the person subject the headline relates to, \texttt{Taggerout Bias} represents the tagged word from the sentence that could potentially be biased, \texttt{Taggerout Prob} refers to the probability of the tagged word been biased, and \texttt{Taggerout in lexicon} refers to the results of checking whether the corresponding tagged word is observed in the lexicon of associated epistemic biased words. This offers some explanability for the potentially biased word. A \textbf{True} value for this column indicates that the word is indeed in the lexicon, whilst a \textbf{False} value indicates otherwise. \texttt{Bias Type} indicates what kind of epistemic bias type is associated with the tagged word. A \textbf{regular} bias type word indicates that the tagged word was not found in the epistemic bias lexicon. The \texttt{Sentence Sentiment} columns represents the overall sentiment associated with the headline sentence (which can be obtained using any NLP processing tool e.g., spacy).


\begin{table*}[htbp]
 \centering
 \begin{adjustbox}{max width=1\textwidth}
 \begin{tabular}{p{0.03\linewidth} | p{0.08\linewidth} | p{0.15\linewidth} | p{0.1\linewidth} | p{0.09\linewidth} | p{0.4\linewidth}| p{0.11\linewidth} }
 \toprule
 No. & Subject & Taggerout Bias &  Taggerout Prob &  Taggerout in Lexicon & Bias Type & Sentence Sentiment \\
 \midrule
 1 & Meghan &   ['staggering'] &  0.999498 & True &   ['subjectives'] &  negative \\
 2  & Kate &  ['astonishing'] &  0.999342 & True &   ['positive', 'subjectives'] &   neutral \\
 3  &  Meghan &   ['beloved'] &  0.997946 & True &   ['positive', 'subjectives'] &  negative \\
 4  & Kate & ['gifted'] &  0.877285 & True &   ['positive', 'subjectives'] &   neutral \\
 5  & Kate & ['packed'] &  0.478948 &   False & ['regular'] &   neutral \\
 6  &  Meghan &   ['incredibly'] &  0.998493 & True &   ['positive', 'subjectives'] &  negative \\
 7  &  Meghan & ['normal'] &  0.689049 & True &   ['subjectives'] &   neutral \\
 8  & Kate &  ['confirms'] &  0.997673 & True &  ['entailments', 'report'] &  negative \\
 9  &  Meghan & ['vanity'] &  0.599388 & True &   ['negative', 'subjectives'] &  negative \\
 10 & Kate & ['top'] &  0.933422 & True &   ['entailments', 'positive', 'subjectives'] &   neutral \\
 11 & Kate & ['homegr'] &  0.820782 &   False & ['regular'] &   neutral \\
 12 &  Meghan &  ['royal'] &  0.551708 &   False & ['regular'] &   neutral \\
 13 & Kate & ['favourite'] &  0.838293 &   False & ['regular'] &   neutral \\
 14 &  Meghan & ['ripped'] &  0.890869 & True &   ['negative'] &   neutral \\
 15 & Kate &  ['bunks'] &  0.758531 & True &   ['negative', 'subjectives'] &  negative \\
 16 &  Meghan &  ['casually'] &  0.967407 &   False & ['regular'] &   neutral \\
 17 & Kate &   ['just'] &  0.645452 & True &   ['subjectives'] &  negative \\
 18 &  Meghan & ['markle'] &  0.583964 &   False & ['regular'] &   neutral \\
 19 & Kate & ['daring'] &  0.856008 & True &  ['entailments', 'implicatives', 'positive', 'subjectives'] &  positive \\
 20 &  Meghan &   ['just'] &  0.565576 & True &   ['subjectives'] &   neutral \\
 \bottomrule
 
 \end{tabular}
 \end{adjustbox}
 \captionsetup{justification=centering}
 \caption{Table showing the output of passing each headline through the model that tags the potentially biased word in the given sentence.}
 \label{tab:tagger_output}
\end{table*}

In Table \ref{tab:costar_output}, the \texttt{Stereotype (S)} column refers to the closest potentially associated stereotype. After passing each headline through the CO-STAR and SBF models as discussed in Section \ref{ssec:methodstereotypes}, we semantically ranked the list of potential stereotype to the headline sentence, and then selected the top ranked potential stereotype and stereotype concept - as seen in the column \texttt{Stereotype Concept(SC)}. The \texttt{S Distance } and \texttt{SC distance} columns shows the semantic similarity distance of the stereotype and stereotype concepts to the headline sentence.

\begin{table*}[htbp]
 \centering
 \begin{adjustbox}{max width=1\textwidth}
 \begin{tabular}{p{0.02\linewidth} | p{0.4\linewidth} | p{0.1\linewidth} | p{0.32\linewidth} | p{0.12\linewidth} }
 \toprule
 No. & Stereotype (S) & Distance(S) & Stereotype Concept(SC) & Distance(SC) \\
 \midrule
 1  & personal spending habits & 0.3457 & women should spend money on clothes & 0.4914 \\
 2  & women should be dressed like brides & 0.3259 & women are property & 0.2278 \\
 3  & feminism & 0.3663 & sexism & 0.2523 \\
 4  & british women are marginalized for a joke & 0.2083 & pregnancy & 0.3918 \\
 5  & N/A & N/A & N/A & N/A \\
 6  & family law & 0.3244 & arab folks are not protected & 0.2951 \\
 7  &  arab folks should live in a constant state of worry & 0.1618 & racial hierarchy & 0.0999 \\
 8  & gender hierarchy & 0.1178 & women are often subjected to terms like ``babies" and ``pregnant" & 0.2632 \\
 9  & women are sexual objects & 0.2602 & misogyny & 0.2574 \\
 10 & women should wear dresses & 0.3131 & women should wear dresses & 0.3131 \\
 11 & women are property of men & 0.1226 & women are property & 0.1494 \\
 12 & women are sex objects & 0.1676 & women are vulnerable & 0.2495 \\
 13 & sexual abuse & 0.0746 & saudi arabians are pedophiles & 0.1226 \\
 14 & sexual assault & 0.2091 & royalty house guests are paedophiles & 0.4691 \\
 15 & women are sex objects & 0.1637 & women are property & 0.1504 \\
 16 & black folks want to marry rich people & 0.2266 & racial marriage & 0.0869 \\
 17 & women are property & 0.2026 & women are secret business partners & 0.3732 \\
 18 & royalty housekeeping are gold diggers & 0.3741 & royalty house queens are gold diggers & 0.3657 \\
 19 & sexism & 0.1927 & gender hierarchy & 0.2284 \\
 20 & women should dress modestly & 0.3840 & women should wear dresses that cover themselves & 0.4047 \\
 \bottomrule
 \end{tabular}
 \end{adjustbox}
 \captionsetup{justification=centering}
 \caption{Table showing the potential top ranked stereotype and stereotype concept. We passed each sentence through the CO-STAR and SBF model. After which the potential stereotype and stereotype concepts where semantically ranked to the sentence itself }
 \label{tab:costar_output}
\end{table*}

\begin{table}[h]
\fontsize{9pt}{10pt}\selectfont
\centering
\resizebox{0.9\columnwidth}{!}{
\begin{tabular}{|c|l|l|}
\hline
\textbf{Headline}                                                                                                                                                                                                                                 & \multicolumn{1}{c|}{\textbf{Participant’s choices (\# of participants)}}                                                                                                                                                                                                                            & \multicolumn{1}{c|}{\textbf{Framework Component’s choice (confidence)}}                                                                                                                                   \\ \hline
\textit{\begin{tabular}[c]{@{}c@{}}Meghan Markle spent a staggering £38,000 on her clothes \\  for a charity trip\end{tabular}}                                                                                                                   & \begin{tabular}[c]{@{}l@{}}Tagged word: staggering (20)\\ epibias\_type: subjective\\ sentiment: negative (15)\\ stereotype: personal spending habits (18)\\ Injustice Present: Character (14)\\ Alternative word: significant (3)\end{tabular}                                                     & \begin{tabular}[c]{@{}l@{}}Tagged word: staggering (0.999498) \\      epibias\_type: subjective\\      sentiment: negative\\      stereotype: personal spending habits\end{tabular}                       \\ \hline

\textit{\begin{tabular}[c]{@{}c@{}}Kate Middleton's £100,000 Astonishing value of the dress that \\  won a Prince’s heart (and has hung in a wardrobe for eight years)\end{tabular}}                                                              & \begin{tabular}[c]{@{}l@{}}Tagged word: astonishing (20)\\  epibias\_type: subjective\\  sentiment: negative (15)\\  stereotype: personal spending habits (18)\\  Injustice Present: Character (14)\\  Alternative word: significant (3)\end{tabular}                                               & \begin{tabular}[c]{@{}l@{}}Tagged word: astonishing (0.999342) \\      epibias\_type: positively subjective\\      sentiment: neutral\\      stereotype: women should be dressed like brides\end{tabular}  \\ \hline

\textit{\begin{tabular}[c]{@{}c@{}}Meghan Markle's beloved avocado linked to human rights abuse \\  and drought, millennial shame\end{tabular}}                                                                                                   & \begin{tabular}[c]{@{}l@{}}Tagged word: beloved (18)\\  epibias\_type: positively subjective\\  sentiment: negative (19)\\  stereotype: feminist are hypocrites (11)\\  Injustice Present: Character (13)\\  Alternative word: `` " (2)\end{tabular}                                                 & \begin{tabular}[c]{@{}l@{}}Tagged word: beloved (0.997946) \\      epibias\_type: positively subjective\\      sentiment: negative\\      stereotype: feminism\end{tabular}                                \\ \hline

\textit{\begin{tabular}[c]{@{}c@{}}Kate’s morning sickness cure? Prince William gifted with \\  an avocado for pregnant Duchess\end{tabular}}                                                                                                     & \begin{tabular}[c]{@{}l@{}}Tagged word: cure (8), gifted (8)\\  epibias\_type: entailments, positively subjective\\  sentiment: neutral (10)\\  stereotype: British women are marginalized for a joke (11)\\  Injustice Present: Character (8)\\  Alternative word: given (4)\end{tabular}          & \begin{tabular}[c]{@{}l@{}}Tagged word: gifted (0.877285) \\      epibias\_type: positively subjective\\      sentiment: neutral\\      stereotype: British women are marginalized for a joke\end{tabular} \\ \hline

\textit{\begin{tabular}[c]{@{}c@{}}Kate and William ‘packed up the kids’ in search of \\  ‘privacy’ at new Windsor estate\end{tabular}}                                                                                                           & \begin{tabular}[c]{@{}l@{}}Tagged word: privacy (8)\\  epibias\_type: regular\\  sentiment: negative (9)\\  stereotype: children are targets of pedos (17)\\  Injustice Present: Character (11)\\  Alternative word: seclusion (2)\end{tabular}                                                     & \begin{tabular}[c]{@{}l@{}}Tagged word: packed (0.478948) \\      epibias\_type: regular\\      sentiment: neutral\\      stereotype: `` "\end{tabular}                                                     \\ \hline

\textit{\begin{tabular}[c]{@{}c@{}}Prince Harry and Duchess Meghan Think Moving to \\  Canada Will Give Archie a Normal Upbringing\end{tabular}}                                                                                                  & \begin{tabular}[c]{@{}l@{}}Tagged word: normal (15)\\  epibias\_type: subjective\\  sentiment: neutral (10)\\  stereotype: racial hierarchy (15)\\  Injustice Present: Character (9)\\  Alternative word: ordinary (2)\end{tabular}                                                                 & \begin{tabular}[c]{@{}l@{}}Tagged word: normal (0.689049) \\      epibias\_type: subjective\\      sentiment: neutral\\      stereotype: racial hierarchy\end{tabular}                                     \\ \hline

\textit{\begin{tabular}[c]{@{}c@{}}Not long to go! Pregnant Kate tenderly cradles her baby bump \\  while wrapping up her royal duties ahead of maternity \\  leave and William confirms she’s due 'any minute now'\end{tabular}}                 & \begin{tabular}[c]{@{}l@{}}Tagged word: tenderly (16)\\  epibias\_type: positively subjective\\  sentiment: positive (13)\\  stereotype: gender hierarchy (15)\\  Injustice Present: none (10)\\  Alternative word: `` " (2)\end{tabular}                                                            & \begin{tabular}[c]{@{}l@{}}Tagged word: confirms (0.997673) \\      epibias\_type: entailments, report\\      sentiment: negative\\      stereotype: gender hierarchy\end{tabular}                         \\ \hline

\textit{\begin{tabular}[c]{@{}c@{}}Why can’t Meghan Markle keep her hands off her bump? \\  Experts tackle the question that has got the nation talking: \\  Is it pride, vanity, acting - or a new age bonding technique?\end{tabular}}          & \begin{tabular}[c]{@{}l@{}}Tagged word: can't (9)\\  epibias\_type: implicative\\  sentiment: negative (13)\\  stereotype: racial hierarchy (10)\\  Injustice Present: Character (19)\\  Alternative word: `` " (2)\end{tabular}                                                                     & \begin{tabular}[c]{@{}l@{}}Tagged word: vanity (0.599388) \\      epibias\_type: negatively subjective\\      sentiment: negative\\      stereotype: women are sex objects\end{tabular}                    \\ \hline

\textit{\begin{tabular}[c]{@{}c@{}}Royal wedding: How Meghan Markle's flowers may have \\  put Princess Charlottes life at risk\end{tabular}}                                                                                                     & \begin{tabular}[c]{@{}l@{}}Tagged word: risk (14)\\  epibias\_type: entailments, negatively subjective\\  sentiment: negative (16)\\  stereotype: relationships (13)\\  Injustice Present: Character (10)\\  Alternative word: could (4)\end{tabular}                                               & \begin{tabular}[c]{@{}l@{}}Tagged word: royal (0.551708) \\      epibias\_type: regular\\      sentiment: neutral\\      stereotype: women are vulnerable\end{tabular}                                     \\ \hline

\textit{\begin{tabular}[c]{@{}c@{}}Kate Middleton’s homegrown bouquet of lily of \\  the valley follows royal code\end{tabular}}                                                                                                                  & \begin{tabular}[c]{@{}l@{}}Tagged word: homegrown (6)\\  epibias\_type: regular\\  sentiment: negative (8)\\  stereotype:gender hierarchy (15)\\  Injustice Present: none (10)\\  Alternative word: `` " (3)\end{tabular}                                                                            & \begin{tabular}[c]{@{}l@{}}Tagged word: homegr (0.820782) \\      epibias\_type: regular\\      sentiment: neutral\\      stereotype: women are property\end{tabular}                                      \\ \hline
\textit{Duchess Kate reveals her favourite photo of son Prince Louis}                                                                                                                                                                             & \begin{tabular}[c]{@{}l@{}}Tagged word: favourite (11), reveals (11)\\  epibias\_type: entailments, factives, report, subjective\\  sentiment: positive (8), neutral (8)\\  stereotype: personal spending habits (18)\\  Injustice Present: none (10)\\  Alternative word:  shares (5)\end{tabular} & \begin{tabular}[c]{@{}l@{}}Tagged word: favourite (0.838293) \\      epibias\_type: regular\\      sentiment: neutral\\      stereotype: saudi arabians are pedophiles\end{tabular}                        \\ \hline

\textit{\begin{tabular}[c]{@{}c@{}}How Meghan and Harry Ripped Up Royal Tradition \\  on Birthday Photos of Archie\end{tabular}}                                                                                                                  & \begin{tabular}[c]{@{}l@{}}Tagged word: ripped (18)\\  epibias\_type: negatively subjective\\  sentiment: negative (15)\\  stereotype: women are sex objects (7)\\  Injustice Present: Character (11)\\  Alternative word: altered (2)\end{tabular}                                                 & \begin{tabular}[c]{@{}l@{}}Tagged word: ripped (0.890869) \\      epibias\_type: negatively subjective\\      sentiment: neutral\\      stereotype: royalty house guests are paedophiles\end{tabular}      \\ \hline

\textit{\begin{tabular}[c]{@{}c@{}}Kate Middleton Debuts New Sapphire Earrings \\  That Belonged to Princess Diana and Debunks a Rumor!\end{tabular}}                                                                                             & \begin{tabular}[c]{@{}l@{}}Tagged word: debunks (10)\\  epibias\_type: refular\\  sentiment: negative (9)\\  stereotype: sexism (10)\\  Injustice Present: none (6)\\  Alternative word: ``remove debunks rumor" (3)\end{tabular}                                                                    & \begin{tabular}[c]{@{}l@{}}Tagged word: bunks (0.758531) \\      epibias\_type: negatively subjective\\      sentiment: negative\\      stereotype: women are sex objects\end{tabular}                     \\ \hline

\textit{\begin{tabular}[c]{@{}c@{}}Meghan Markle Just Casually Rewore Her \$16,500 \\  Royal Wedding Earrings in NYC\end{tabular}}                                                                                                                & \begin{tabular}[c]{@{}l@{}}Tagged word: casually (16)\\  epibias\_type: regular\\  sentiment: negative (15)\\  stereotype: black folks want to marry rich people (8)\\  Injustice Present: Character (11)\\  Alternative word: ``remove casually) + wore (7)\end{tabular}                            & \begin{tabular}[c]{@{}l@{}}Tagged word: markle (0.583964) \\      epibias\_type: regular\\      sentiment: neutral\\      stereotype:\end{tabular}                                                         \\ \hline

\textit{\begin{tabular}[c]{@{}c@{}}A right royal cash in! How Prince Harry and Meghan Markle \\  trademarked over 100 items from hoodies to socks \\  SIX MONTHS before split with monarchy - with new \\  empire worth up to £400m\end{tabular}} & \begin{tabular}[c]{@{}l@{}}Tagged word: cash in (13)\\  epibias\_type: regular\\  sentiment: negative (16)\\  stereotype: royalty housekeeping are gold diggers (8)\\  Injustice Present: Character (15)\\  Alternative word: established (3)\end{tabular}                                          & \begin{tabular}[c]{@{}l@{}}Tagged word: casually (0.967407) \\      epibias\_type: regular\\      sentiment: neutral\\      stereotype: royalty housekeeping are gold diggers\end{tabular}                 \\ \hline

\textit{\begin{tabular}[c]{@{}c@{}}Kate and Wills Inc: Duke and Duchess secretly set up companies \\  to protect their brand - just like the Beckhams\end{tabular}}                                                                               & \begin{tabular}[c]{@{}l@{}}Tagged word: secretly (10)\\  epibias\_type: regular\\  sentiment: neutral (10)\\  stereotype: women should be protected from being defined by men (12)\\  Injustice Present: Framing(7), Character (7)\\  Alternative word: ``remove secretly" (6)\end{tabular}          & \begin{tabular}[c]{@{}l@{}}Tagged word: just (0.645452) \\      epibias\_type: subjective\\      sentiment: negative\\      stereotype: women are property\end{tabular}                                    \\ \hline
\end{tabular} 
}
\caption{Comparing Survey Results for Participants and Framework Outputs \label{tab:surveysumcomparison}}
\end{table}

\end{document}